\documentclass{article}

\PassOptionsToPackage{numbers, compress}{natbib}

\usepackage[preprint]{neurips_2024}

\usepackage[hidelinks]{hyperref}
\usepackage[hang,flushmargin]{footmisc}

\usepackage[utf8]{inputenc} 
\usepackage[T1]{fontenc}    
\usepackage{hyperref}       
\usepackage{url}            
\usepackage{booktabs}       
\usepackage{amsfonts}       
\usepackage{nicefrac}       
\usepackage{microtype}      
\usepackage{xcolor}         
\usepackage{amsmath}
\usepackage{enumitem}
\usepackage{graphicx}       
\usepackage{caption}
\usepackage{subcaption}
\usepackage{threeparttable}
\usepackage{color, colortbl}
\usepackage{threeparttable,booktabs}
\usepackage{multirow}
\usepackage{balance}
\usepackage{xspace}
\usepackage{balance}
\usepackage{soul} 
\usepackage{wasysym}
\usepackage{xcolor,colortbl}
\usepackage{mdframed}
\usepackage{floatrow}
\PassOptionsToPackage{dvipsnames}{xcolor}
\usepackage{ragged2e}

\newcommand\ignore[1]{}

\newcommand{\hlgreen}[1]{\sethlcolor{green!30}\hl{#1}}
\newcommand{\hlorange}[1]{\sethlcolor{orange!30}\hl{#1}}
\newcommand{\hlred}[1]{\sethlcolor{red!15}\hl{#1}}
\newcommand{\trecrag}{TREC 2024 RAG Track\xspace}
\newcommand{\msmarco}{MS MARCO V2.1 segment collection\xspace}
\newcommand{\vmsmarco}{MS MARCO V2 document collection\xspace}

\definecolor{dark_green}{rgb}{0.0, 0.5, 0.0}

\title{Support Evaluation for the TREC 2024 RAG Track: Comparing Human versus LLM Judges}

\author{%
  Nandan Thakur$^1$, Ronak Pradeep$^1$, Shivani Upadhyay$^1$, \\ \textbf{Daniel Campos}$^2$, \textbf{Nick Craswell}$^3$, \textbf{Jimmy Lin}$^1$ \\
  \\
  $^1$~University of Waterloo \quad $^2$~Snowflake \quad $^3$~Microsoft \\
  \\
  \url{https://trec-rag.github.io}
}

\begin{document}

\maketitle

\begin{abstract}
Retrieval-augmented generation (RAG) enables large language models (LLMs) to generate answers with citations from source documents containing ``ground truth'', thereby reducing system hallucinations. 
A crucial factor in RAG evaluation is ``support''---whether the information in the cited documents supports the answer.
To this end, we conducted a large-scale comparative study of 45 participant submissions on 36 topics to the \trecrag, comparing an automatic LLM judge (GPT-4o) against human judges for support assessment. 
We considered two conditions:\ (1) fully manual assessments from scratch and (2) manual assessments with post-editing of LLM predictions. 
Our results indicate that for 56\% of the manual from-scratch assessments, human and GPT-4o predictions match perfectly (on a three-level scale), increasing to 72\% in the manual with post-editing condition.
Furthermore, by carefully analyzing the disagreements in an unbiased study, we found that an independent human judge correlates better with GPT-4o than a human judge, suggesting that LLM judges can be a reliable alternative for support assessment. 
To conclude, we provide a qualitative analysis of human and GPT-4o errors to help guide future iterations of support assessment. 
\end{abstract}

\section{Introduction}

Retrieval-Augmented Generation (RAG) has recently gained popularity in both academic and industrial settings (e.g., Bing Search \cite{bingsearch} and popular frameworks like LangChain \cite{langchain}).
In RAG, large language models (LLMs) generate answers to user queries that include citations to source documents as necessary \cite{guu:2020,lewis:2020,izacard:2021,borgeaud:2022}.
RAG systems improve factuality and verifiability, reducing hallucinations observed in ``closed-book'' LLM generation \cite{khandelwal:2020,lewis:2020,gao:2023,liu:2024}. 

Document-level citations for supporting facts in LLM-generated answers are integral to any deployed RAG system.
Therefore, support evaluation assesses whether a RAG answer factually supports the information present in the cited documents, which is crucial for evaluating the quality of a RAG system.
Prior work on support evaluation in the RAG literature \cite{liu-etal-2023-evaluating, Chen_Lin_Han_Sun_2024, saad-falcon-etal-2024-ares, es-etal-2024-ragas,wu:2024,yu:2024, ru:2024} relies on an \emph{automatic} judge, i.e., an LLM as a proxy judge. 
However, it is unknown whether an LLM judge can potentially replace a human judge for support evaluation.

This paper examines results from the \trecrag, assessing 45 participant systems on 36 information-based queries.
A sample query and answer is shown in \autoref{tab:taylor-swift-answer}.
We conducted a large-scale comparative study between human and LLM judges using resources provided by the National Institute of Standards and Technology (NIST) to better understand whether support assessment can be automated.
Unique to the TREC setup, we contrast our automatic judgment process using a strong LLM judge (like GPT-4o) against a manual process under two conditions:\ (1) \emph{manual from scratch}, where human annotators perform assessments from scratch and (2) \emph{manual with post-editing}, where human annotators are shown GPT-4o predictions during the evaluation process.

In this paper, we focus exclusively on support, i.e., whether the information in an answer sentence is supported by the cited documents, which we consider as the ``ground truth''.
This is, of course, only one aspect of RAG evaluation.
For in-depth analyses of other aspects of the \trecrag, we refer readers to \citet{upadhyay:2024} for relevance assessment and \citet{pradeep:2024} for nugget evaluation.

Our experimental results indicate that GPT-4o and human judgments perfectly match 56\% of the time in the manual from-scratch condition, increasing to 72\% in the manual with post-editing condition.
These results show promise in using LLM judges for support assessment in both conditions.
We measured support of a system's overall answer in terms of two metrics:\ weighted precision and weighted recall, where precision penalizes overcitation, and recall penalizes undercitation.
We observe a high correlation at the run level (above 0.79 Kendall $\tau$) between GPT-4o and human judges, providing evidence that LLMs can potentially replace human judges for support evaluation.

In addition, to better understand the discrepancies between GPT-4o and human judges, we conducted an unbiased disagreement study with an independent human judge who carefully re-assessed 537 randomly sampled pairs, including both assessment conditions.
Our results surprisingly show that the independent judge agrees more with GPT-4o than the human judge (e.g., Cohen's $\kappa$ of 0.27 vs.~0.07).
Finally, we discuss annotation errors to help improve future iterations of support evaluation. 

\begin{table}[t]
\setlength{\tabcolsep}{3pt}

\renewcommand{\arraystretch}{1}
\centering
\resizebox{\textwidth}{!}{
{\begin{tabular}{p{15cm}}
\toprule
{\bf Topic: how taylor swift's age affects her relationships} \\[1ex]
{\bf Answer}: Taylor Swift's age has significantly influenced her relationships, often contributing to both the dynamics and outcomes of her romances \hl{[ ]}.
Swift's relationships have frequently involved notable age gaps, which have sometimes led to complications \hl{[ ]}.
For instance, her relationship with John Mayer, who was 11 years her senior, reportedly strained due to the age difference and Mayer's reputation as a playboy, leading to Swift feeling taken advantage of \hl{[0, 3, 4]}.
This relationship inspired her song "Dear John," reflecting her emotional turmoil \hl{[3, 4]}.
Similarly, Swift's relationship with Jake Gyllenhaal, who was nine years older, also highlighted the impact of age differences.
Gyllenhaal reportedly felt the age gap, and Swift's inexperience made it difficult for her to navigate the complexities of the relationship, leading to heartbreak and inspiring several songs on her "Red" album \hl{[8, 1, 2]}. [...] 
In summary, Taylor Swift's age has played a crucial role in shaping her relationships, influencing both the dynamics and outcomes \hl{[8, 1]}.
Age differences have often led to complications, while relationships with peers have been relatively simpler.
Over time, her maturity has allowed her to navigate romantic complexities with greater understanding and resilience. \hl{[ ]} \\ \midrule
\textbf{Passage [3]:} A Timeline Of Taylor Swift's Age-Inappropriate Romances | Business Insider [...] Taylor Swift, 21, \& John Mayer, 32. The inappropriateness of Swift's dating habits peaked when she started seeing 11-years her senior and notorious playboy, John Mayer. Neither party confirmed the romance, but they reportedly dated through February 2010 [...] \\[-0.5em] \textcolor{lightgray}{\rule[-0.1ex]{\linewidth}{0.08em}}
\textbf{Passage [8]:} What Really Went Wrong With Jake Gyllenhaal And Taylor Swift? [...] Taylor Swift and Jake Gyllenhaal dated from October to December 2010. [...] What went wrong with these two ``\textit{twin fire signs?}'' It turns out, a lot. Maybe that's why it's such a touchy subject for him these days! The age gap, when they dated, Gyllenhaal was 29 to Swift's 20, and by the time they split, he was 30 to her 21. ``\textit{He said he could feel the age difference},'' a source told Us Weekly [...] \\
\bottomrule
\end{tabular}}}
\caption{A sample RAG answer segmented into multiple sentences with citations to passages retrieved from the \msmarco for a query about Taylor Swift. We \hl{highlight} the cited passages, which are provided as a list (that can be empty, i.e., zero citations). }
\label{tab:taylor-swift-answer}
\end{table}

\section{Background and Related Work}

What we refer to as support evaluation represents an attempt to determine whether information in a RAG answer is factually supported by its cited documents.
In our work, we evaluate support at the sentence level in the answer.
We assume an answer $r$ segmented into $n$ sentences, $r = \{a_{1},\cdots,a_{n}\}$, where each answer sentence $a_i$ can contain a maximum of $m$ document citations, $a_i = \{d_i, \cdots,d_m\}$, each of which are documents drawn from a corpus.\footnote{Consistent with parlance in the field, we refer to documents in the generic sense, even though in actuality a ``document'' may be a passage (as in our case), a PDF, or even an image.}
Support is calculated as the function $f(a_i, d_j)=s_{i,j}$ where $f$ can be a human or LLM judge that generates a scalar value $s_{i,j}$, indicating the extent that the cited document $d_j$ provides support to sentence $a_i$.
A few examples of support evaluation are shown in \autoref{tab:sample-retrieval}.
Apart from RAG, support has been primarily explored in the literature for summarization \cite{laban:2023, jia:2023} and natural language explanations \cite{atanasova:2023, siegel:2024}.

\begin{table}[t]
\setlength{\tabcolsep}{3pt}

\renewcommand{\arraystretch}{1}
\centering
\resizebox{0.5\textwidth}{!}{
{\begin{tabular}{p{8cm}}
\toprule
{\bf Answer Sentence:} For instance, her relationship with John Mayer, who was 11 years her senior, reportedly strained due to the age difference and Mayer's reputation as a playboy, leading to Swift feeling taken advantage of. \\ \midrule
{\bf Passage ID [0]:} \texttt{doc\_04\_1081579649\#7\_2253255175} \\
{\bf Title:} Timeline of Taylor Swift's Relationships \\
{\bf Text:} 2009: Taylor Swift, 20, \& Taylor Lautner, [...] 2010: Taylor Swift, 21, \& John Mayer, 321 / 12 And then the \hl{inappropriateness of Swift's dating habits peaked when she started seeing 11-years her senior and notorious playboy, John Mayer.} \\[-0.5em] \textcolor{lightgray}{\rule[-0.1ex]{\linewidth}{0.08em}}
{\bf Human Judge:} \hlorange{Partial Support} \\ {\bf GPT-4o Judge:} \hlorange{Partial Support} \\ \midrule
\midrule
{\bf Answer Sentence:} This relationship inspired her song ``Dear John,'' reflecting her emotional turmoil. \\ \midrule
{\bf Passage ID [3]:} \texttt{doc\_35\_202251892\#8\_427548986} \\
{\bf Title:} Timeline Of Taylor Swift's Age-Inappropriate Romances | Business Insider \\
{\bf Text:} [...] 2010: Taylor Swift, 21, \& John Mayer, 321 / 12 And then the inappropriateness of Swift's dating habits peaked  [...]  and \hl{then the heartbroken young Swift penned the song 'Dear John' about the break up.} Earlier this year, Mayer admitted that he felt `humiliated' when he heard the song, but Swift refuses to admit it's about him, telling Glamour magazine it was `presumptuous' of him to think the song was about him. \\[-0.5em] \textcolor{lightgray}{\rule[-0.1ex]{\linewidth}{0.08em}}
{\bf Human Judge:} \hlgreen{Full Support} \\ {\bf GPT-4o Judge:} \hlgreen{Full Support}  \\
\bottomrule
\end{tabular}}}
\hspace{0.1cm}
\resizebox{0.48\textwidth}{!}{
{\begin{tabular}{p{8cm}}
\toprule
{\bf Answer Sentence:} The age difference was a significant factor in their breakup, with Gyllenhaal not ready to commit, further exacerbating Swift's emotional distress. Conversely, Swift's relationships with peers closer to her age, such as Joe Jonas, were less fraught with such issues. \\ \midrule
{\bf Passage ID [8]:} \texttt{doc\_48\_737500982\#1\_1325021022} \\
{\bf Title:} What Went Wrong With Jake Gyllenhaal And Taylor Swift? \\
{\bf Text:} Taylor Swift and Jake Gyllenhaal dated from October to December 2010. [...] ``He said \hl{he could feel the age difference,}'' a source told Us Weekly. [...] ``When Jake broke her heart, \hl{she was so inexperienced she didn't know how to deal with it ... She wasn't used to all the head games and the lies} but now she's way less naive.'' \\[-0.5em] \textcolor{lightgray}{\rule[-0.1ex]{\linewidth}{0.08em}}
{\bf Human Judge:} \hlgreen{Full Support} \\ {\bf GPT-4o Judge:} \hlorange{Partial Support} \\
\midrule \midrule
{\bf Answer Sentence:} As she matured, her understanding of relationships evolved, making her less naive and more discerning in her romantic choices. \\ \midrule
{\bf Passage ID [8]:} \texttt{doc\_48\_737500982\#1\_1325021022} \\
{\bf Title:} What Went Wrong With Jake Gyllenhaal And Taylor Swift? \\
{\bf Text:} Taylor Swift and Jake Gyllenhaal dated from October to December 2010. [...] ``He said he could feel the age difference,'' a source told Us Weekly. [...] ``When Jake broke her heart, she was so inexperienced she didn't know how to deal with it. She wasn't used to all the head games and the lies but now she's way less naive.'' \\[-0.5em] \textcolor{lightgray}{\rule[-0.1ex]{\linewidth}{0.08em}}
{\bf Human Judge:} \hlred{No Support} \\ {\bf GPT-4o Judge:} \hlorange{Partial Support} \\
\bottomrule
\end{tabular}}}
\caption{Examples of support evaluation with GPT-4o and human judges for the Taylor Swift topic ``\textit{how taylor swift's age affects her relationships}''. The fragment of the passage that supports the answer sentence is \hl{highlighted}.}
\label{tab:sample-retrieval}
\end{table}

Previous work on support evaluation in RAG used different automatic judges:\
examples include an natural language inference (NLI) model \cite{gao:2023}, LLM with prompting \cite{es-etal-2024-ragas}, or even fine-tuned custom LLMs \cite{saad-falcon-etal-2024-ares} as the automatic judge.
\citet{wu:2024} evaluated the tug of war between an LLM's internal prior over supporting wrong context information.
Similar to our formulation, \citet{ming:2024} provided an evaluation benchmark consisting of academic question answering (QA) datasets with human validation and \citet{liu-etal-2023-evaluating} evaluated the quality of proprietary search engine outputs with crowdsourced human judges. 
In contrast, our work is one of the first to conduct a large-scale human annotation study---encompassing 11K human assessments over multiple RAG systems on 36 topics containing non-factoid, decompositional, and multi-perspective queries.
This study design provides a rich backdrop for comparing human and GPT-4o judges for support evaluation. 

\section{Track Description \& Assessment Methodology}

\subsection{\trecrag}

The Text Retrieval Conference (TREC) has led the way in many aspects of evaluation in information retrieval (IR), natural language processing (NLP), and beyond, for accelerating research within the community (both researchers and practitioners). 
Each year, TREC organizes several tracks, focused on topics ranging from text or multimodal retrieval \cite{craswell2022overview, Craswell_etal_TREC2023, yang:2023} to conversational QA \cite{aliannejadi:2024}.

The context for this work is the \trecrag,
which was divided into three tasks:\ retrieval (R), augmented generation (AG), and retrieval-augmented generation (RAG).
Here, we focus on the generation part, i.e., participant systems are presented with queries (called topics in TREC parlance) and candidate passages.
These candidate passages are either generated by us (i.e., the track organizers) and shared with all the participants~\cite{ragnarok} (the AG task) or each participant can directly perform retrieval from the \msmarco (the end-to-end RAG task).
The candidate passages provide the context or grounding of RAG for synthesizing the final free-form answer.
We require that answers be segmented into sentences, and that each sentence is associated with citations to passages from the corpus, as shown in \autoref{tab:taylor-swift-answer}.
Since many teams participated in the \trecrag, our human and LLM judges were exposed to multiple answers and cited documents during support evaluation.

\paragraph{Passage collection.}
The \msmarco contains 113,520,750 text passages, derived from a deduplicated version of the \vmsmarco~\cite{craswell2022overview} by removing near-duplicate documents using locality-sensitive hashing (LSH) with MinHash and 9-gram shingles.
This reduced the original document count from 11,959,635 to 10,960,555 documents. 
Passages were derived from the corresponding document collection using a sliding-window chunking technique with overlap---specifically, using windows of 10 sentences with a stride of 5 sentences, producing passages typically between 500--1000 characters.
Each passage comprises a \texttt{title} field containing the title of the passage and a \texttt{text} field containing the body of the passage.

\paragraph{Topic collection.}
For the \trecrag topics (queries), we leveraged a fresh scrape of Bing Search logs containing non-factoid queries that are multifaceted and subjective, warranting RAG systems to provide long-form answers~\cite{ragnarok, Rosset:2402.17896:2024}. 
We gathered topics close to the submission period of the evaluation (around July 2024) to avoid staleness and minimize potential data leakage. 
Here, we are not concerned about data contamination in support assessment:\ since we are retrieving from web corpora, our passages are likely present in LLM pre-training data. 
Due to budget constraints with human annotations, we only ran evaluations using a subset of 36 topics selected from the complete \trecrag topic collection.

\subsection{Support Assessment}

Consistent with previous support evaluations in RAG~\cite{liu-etal-2023-evaluating, gao:2023}, we used a three-level grade, with the following associated descriptions for each support level:

\begin{itemize}[leftmargin=0.75cm]

\item [\textbf{FS}] \hlgreen{\textbf{Full Support}}: All of the information in the answer sentence is factually consistent with and supported by the cited passage.

\item [\textbf{PS}] \hlorange{\textbf{Partial Support}}: Some of the information in the answer sentence is factually consistent with and supported by the cited passage, but other parts of the sentence are not supported.

\item [\textbf{NS}] \hlred{\textbf{No Support}}: The cited passage is completely irrelevant and does not support any part of the answer sentence.

\end{itemize}

An edge case is a sentence with zero citations:\
We automatically consider the support assessment to be ``no support'', as the sentence does not cite any retrieved passage.

Next, in order to evaluate the quality of LLM judges in contrast to human judges, we conducted our support assessment with human judges under two conditions:\ (1) manual from scratch and (2) manual with post-editing.
We describe both conditions in detail below:

\begin{enumerate}[leftmargin=0.75cm]

\item {\bf Manual from scratch.}
In this condition, a human judge is provided with the answer sentence and the cited passage.
The judge reads both carefully and evaluates whether the answer sentence is supported by the cited passage (according to one of the labels above).

\item {\bf Manual with post-editing.}
In this condition, a human judge is provided with the answer sentence, the cited passage, and the support judgment label given by the LLM judge.
The human judge reads the sentence and passage carefully and provides an assessment using the LLM judgment label as a reference.

\end{enumerate}

\noindent For automatic labeling, we utilized GPT-4o as an automatic judge.
We ran inference using the Microsoft Azure API \cite{openai_gpt4o}, providing a single passage at a time in the prompt\footnote{Separately, we experimented with providing multiple cited passages at once, but we anecdotally observed that providing one passage at a time performs better with the GPT-4o judge.} using the answer sentence and the cited passage.
The GPT-4o judge is presented with each sentence and its cited passage and asked to determine the support label without any explanation (full support, partial support, or no support). 
The prompt used is given in \autoref{fig:prompt}.

\begin{figure}[t]
\begin{mdframed}[font=\small, roundcorner=10pt, linecolor=purple, linewidth=1pt, innerleftmargin=10pt, innerrightmargin=10pt, innertopmargin=10pt, innerbottommargin=10pt]
In this task, you will evaluate whether each statement is supported by its corresponding citations. Note that the system responses may appear very fluent and well-formed, but contain slight inaccuracies that are not easy to discern at first glance. Pay close attention to the text. \\

You will be provided with a statement and its corresponding passage which the statement cites. It may be helpful to ask yourself whether it is accurate to say ``\emph{according to the citation ...}'' with the statement following this phrase. Be sure to check all of the information in the statement. You will be given three options: \\

• Full Support: All of the information in the statement is supported in the citation.

• Partial Support: Some parts of the information are supported in the citation, but other parts are missing.

• No Support: The citation does not support any part of the statement. \\

Please provide your response based on the information in the citation. If you are unsure, use your best judgment. Respond as either ``Full Support'', ``Partial Support'', or ``No Support'' with no additional information.

Statement: \{$statement$\}

Citation: \{$citation$\}
\end{mdframed}
\vspace{-2mm}
\caption{Prompt used by the GPT-4o judge for support evaluation.}
\label{fig:prompt}
\end{figure}

\subsection{Computational Cost \& Evaluation Tradeoffs}

In the \trecrag, we allowed participants to provide citations for up to 20 passages per answer sentence.
To judge each sentence and its cited passage, our protocol requires a human judge to read the answer sentence and a relatively long text passage (typically, 500--1000 characters). 
Thus, conducting an exhaustive evaluation of all cited passages for every answer sentence across multiple participants was not feasible given our budget constraints.

Therefore, we had to choose between sparse and dense annotations.
Dense annotations would provide fewer judged topics, but each answer sentence would be evaluated against {\it k} cited passages.
On the other hand, sparse annotations would provide higher diversity in judged topics, but at the cost of judging fewer cited passages for every answer sentence. 

We opted for sparse annotations to achieve more judged topics.
We fixed both the human and GPT-4o judge to evaluate only the first cited passage of every answer sentence for all participants.
As with all TREC evaluations, NIST provided the resources to perform human evaluations based on the guidance of the track organizers (i.e., us). 
NIST first trained every human judge to understand the task, and then each human judge evaluated each topic sequentially.

\subsection{Support Evaluation Metrics}

Support can be evaluated across two dimensions, similar to Liu et al. \cite{liu-etal-2023-evaluating}:\ (1) {\it weighted precision}, accounting for how many correct passage citations are present in the generated answer, and (2) {\it weighted recall}, accounting for how many sentences in the answer are supported by passage citations
We define both metrics below:

\paragraph{Weighted precision.}
This metric measures the weighted proportion of citations that support each answer sentence.
We assign a weight to $s(a_i, d_j)$ of $1.0$ to Full Support (FS), $0.5$ to Partial Support (PS), and $0$ to No Support (NS) for the answer sentence and cited passage. 
To explain the metric clearly, let us assume a RAG answer with $3$ sentences = $\{a_1, a_2, a_3\}$, and a corpus $C$ with $2$ passages:\ $\{p_1, p_2\}$. 
Now, let's assume that passage $p_1$ partially supports $a_1$, passage $p_2$ fully supports $a_2$, and $a_3$ has zero citations. 
We compute weighted precision as follows:
\begin{equation*}
\text{Weighted Precision} = \dfrac{s(a_1,p_1) + s(a_2,p_2)}{\text{count}(\{a_1,p_1\},\{a_2,p_2\})} = \dfrac{0.5 + 1}{2} = 0.75
\end{equation*}

\begin{table}[t]
  \resizebox{\columnwidth}{!}{%
  \begin{tabular}{ll |  c  c | c c c}
    \toprule
    & \multirow{2}{*}{\textbf{Condition}} & \multirow{2}{*}{\textbf{\#Topics}} & \multirow{2}{*}{\textbf{\#Annotations}} & \multicolumn{3}{c}{\textbf{Support level}} \\
    &&& \textbf{} & \textbf{FS} & \textbf{PS} & \textbf{NS}\\
    \midrule
    (1a) & Manual from scratch (Human) & 22  & 6,742 & 2,752 & 1,652 & 2,338 \\
    (1b) & Automatic (GPT-4o) & 22 & 6,742 & 3,110 & 2,421 & 1,211 \\ \midrule
    (2a) & Manual with post-editing (Human) & 14 & 4,165 & 1,812 & 1,076 & 1,277 \\
    (2b) & Automatic (GPT-4o) & 14 & 4,165 & 2,045 & 1,330 & 790 \\
  \bottomrule
  \end{tabular}
  }
  \vspace{0.1cm}
  \caption{Descriptive statistics for support judgments for the (1) manual from-scratch condition and (2) manual with post-editing condition for 45 participant submissions on 36 topics.}
  \vspace{-0.3cm}
  \label{tab:support_stats}
\end{table}

\paragraph{Weighted recall.}
This metric measures the weighted proportion of answer sentences that are supported by their cited passages. 
We assign the same weights as defined above in weighted precision.
For the above example, we compute weighted recall as follows:

\begin{equation*}
\text{Weighted recall} = \dfrac{s(a_1,p_1) + s(a_2,p_2)}{\text{count}(\{a_1, a_2, a_3\})} = \dfrac{0.5 + 1}{3} = 0.5
\end{equation*}

\noindent Precision penalizes {\it overcitation} in the answer text that is often distracting and unnecessary for the user. 
On the other hand, recall penalizes answers with {\it undercitation}, i.e., sentences with zero citations.
Therefore, if a sentence has zero citations, it reduces the recall score but keeps precision unchanged.
As described above, we evaluated only the first cited passage per answer sentence for support.
Therefore, the weighted recall and precision scores are identical if all answer sentences have at least one citation. 
We save for future work how to best evaluate multiple passage citations for every sentence in the answer, as it would require a much larger annotation budget.

\begin{figure*}[t]
        \centering
        \includegraphics[width=0.495\textwidth]{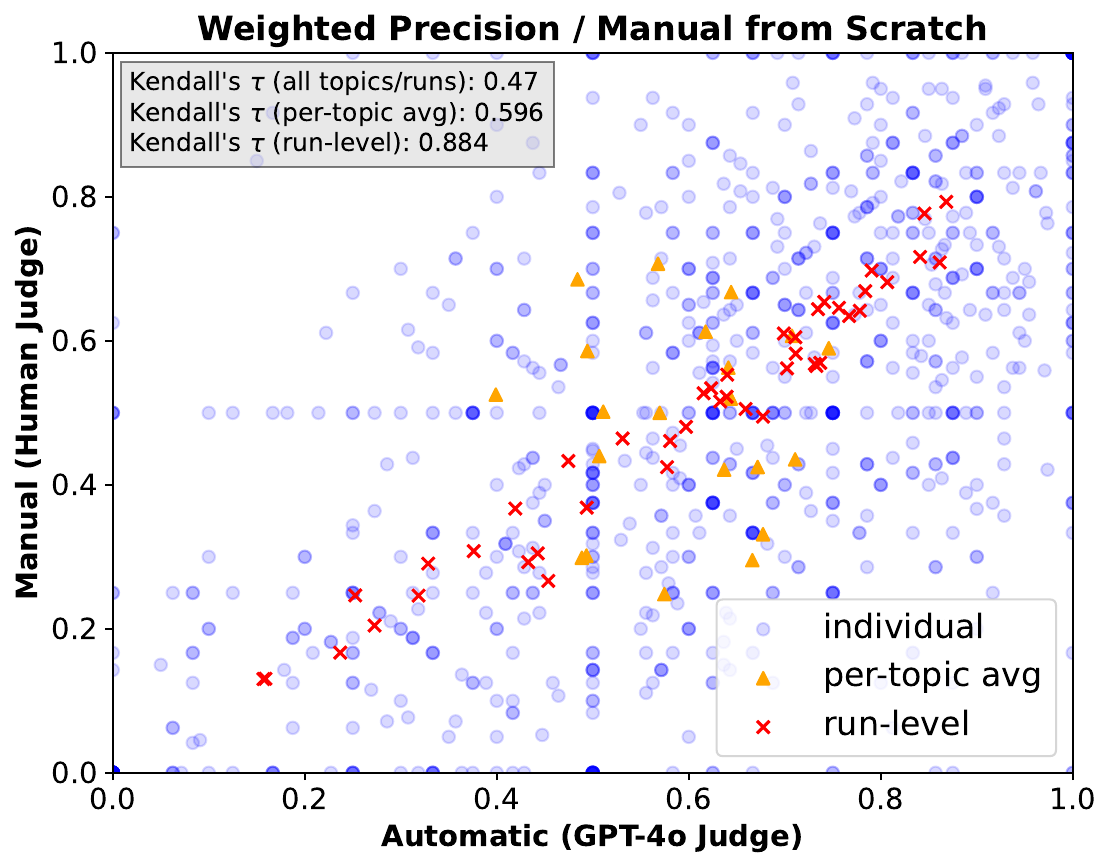}
        \includegraphics[width=0.495\textwidth]{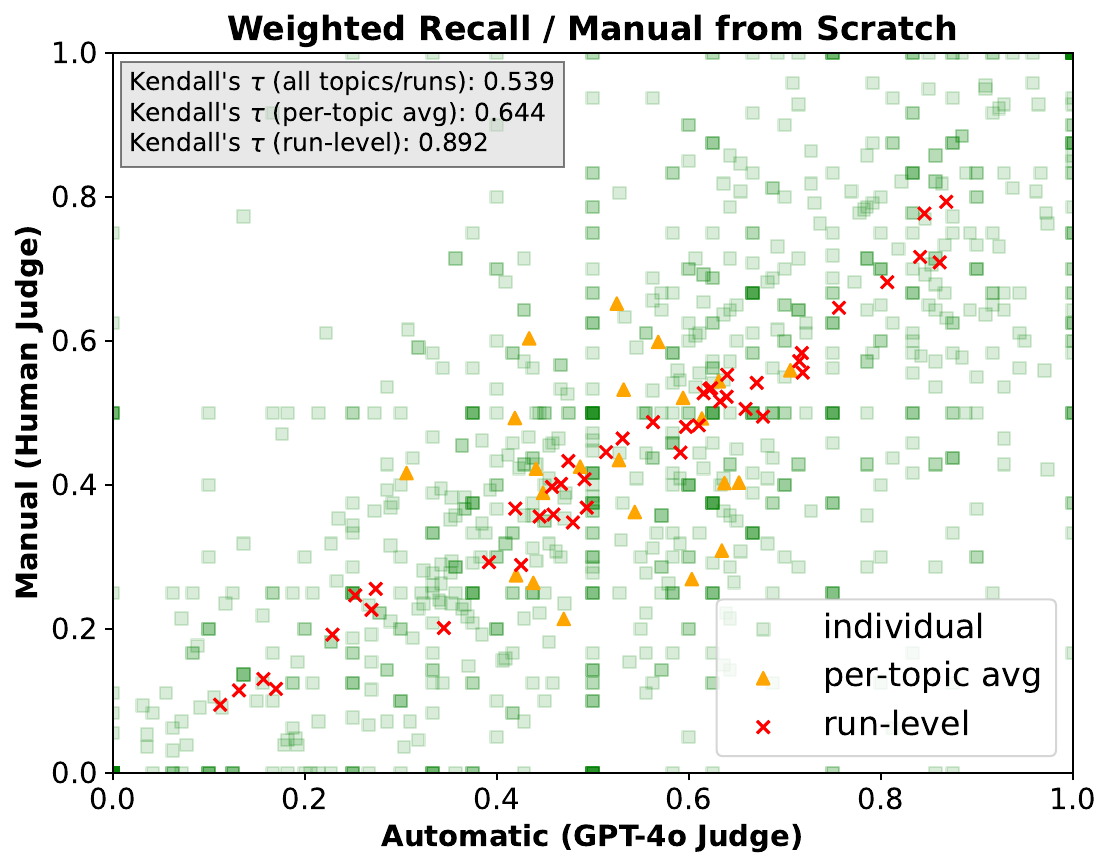}
        \includegraphics[width=0.495\textwidth]{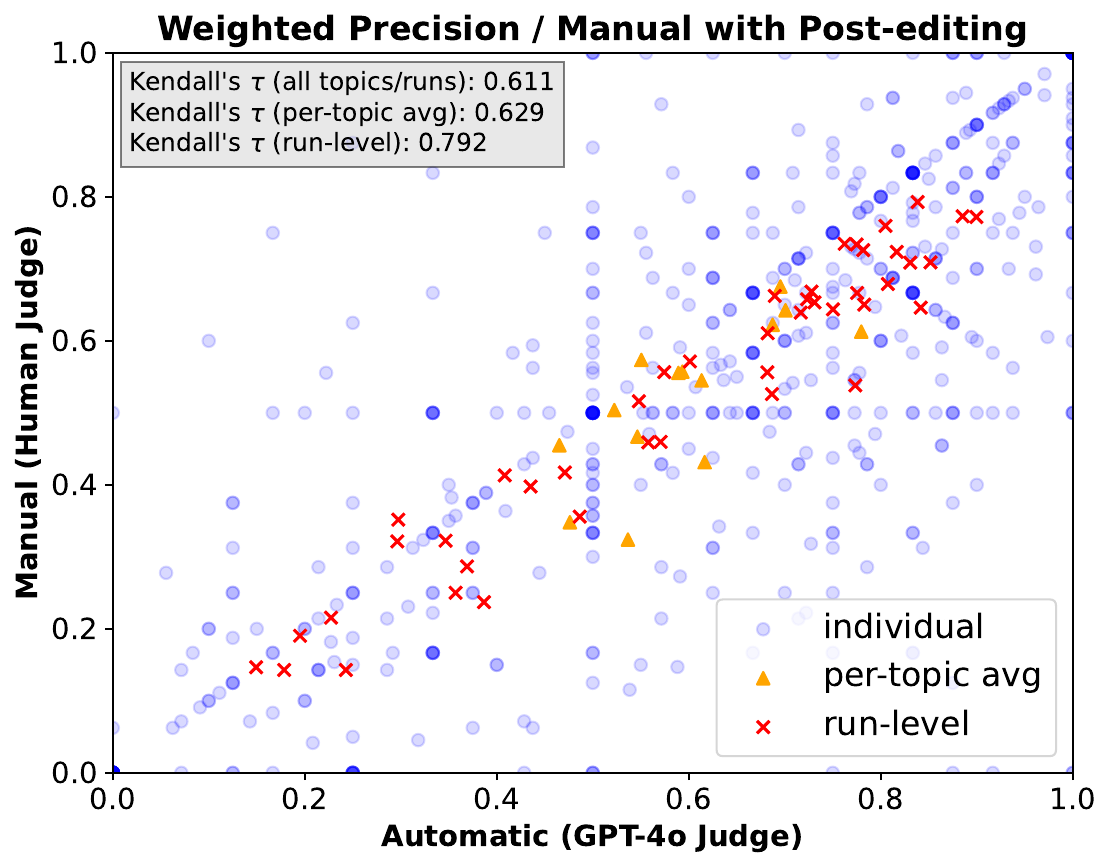}
        \includegraphics[width=0.495\textwidth]{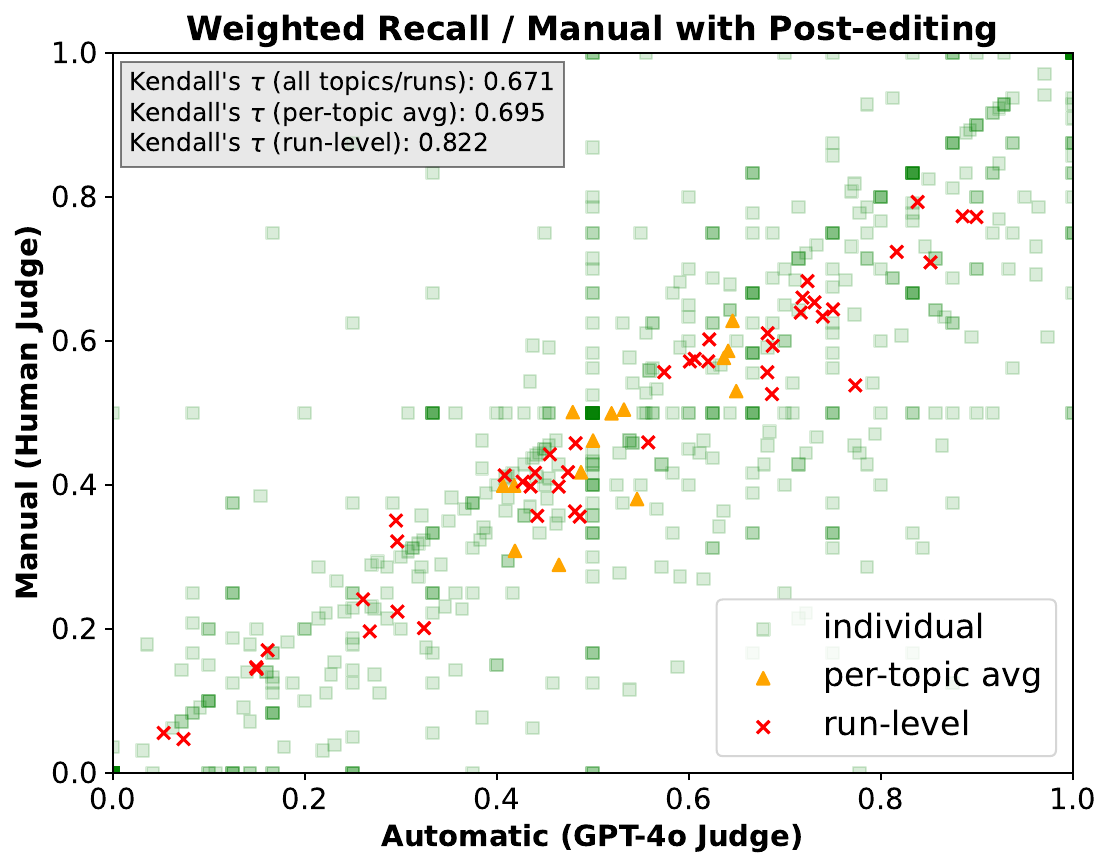}
    \caption{Correlations between scores from human and GPT-4o judges for the manual from-scratch condition (top) and the manual with post-editing condition (bottom), measuring weighted precision and recall. Red markers show run-level scores, yellow triangles show per-topic averages, and blue dots or green boxes show all individual topic/run combinations. Each plot is annotated with rank correlations showing Kendall's $\tau$.}
    \label{fig:scatter_plots_1}
\end{figure*}

\section{Experimental Results}
\label{sec:results}

For the \trecrag, NIST received 93 runs from 20 groups for the end-to-end RAG task and 53 runs from 11 groups for the AG task.
Given resource constraints, we evaluated only the two highest-priority submissions from each group across the RAG and AG tasks. 
As shown in \autoref{tab:support_stats}, this translates into 31 runs from 18 groups for RAG and 14 runs from 9 groups for AG, totaling a maximum of 45 participant submissions for each topic.
The human judges were able to complete judgments for 36 topics, sparsely annotated:\ 6,742 annotations on 22 topics in the manual from-scratch condition and 4,165 annotations on 14 topics in the manual with post-editing condition.

\subsection{Weighted Precision and Recall}

\autoref{tab:support_manual_from_scratch} and \autoref{tab:support_manual_from_scratch_auto} show the average weighted precision and recall scores of all participant runs for both the AG and RAG tasks in the manual from-scratch condition on the 22 topics evaluated by human and GPT-4o judges, respectively. 
\autoref{tab:support_manual_with_post_editing} and \autoref{tab:support_manual_with_post_editing_auto} show the average weighted precision and recall scores for all participants in the manual with post-editing condition on the 14 topics evaluated by human and GPT-4o judges, respectively.
We sort the runs in terms of the average weighted precision score in descending order.

In \autoref{fig:scatter_plots_1}, we show scatter plots of weighted precision and recall scores obtained by all participant submissions.
Run-level scores (denoted by $\times$) are strongly correlated (all above 0.79 Kendall's $\tau$) between GPT-4o and human annotations. 
Per-topic averages (denoted by $\triangle$) vary on both axes, where certain topics achieve a higher weighted precision and recall score than humans over GPT-4o, and vice versa.
Individual participant scores (denoted by $\Circle$ or $\square$) show a high variance in both weighted precision and recall scores.
This is likely due to the mismatch of human annotators preferring ``no support'', whereas GPT-4o prefers ``partial support''. 
Overall, we observe the majority of scores in the bottom right triangle, indicating that humans take a more conservative approach and provide lower levels of support overall than GPT-4o, leading to lower weighted precision and recall scores.

\subsection{Confusion Matrices}

\begin{figure}[t!]
        \centering
        \includegraphics[width=0.45\textwidth]{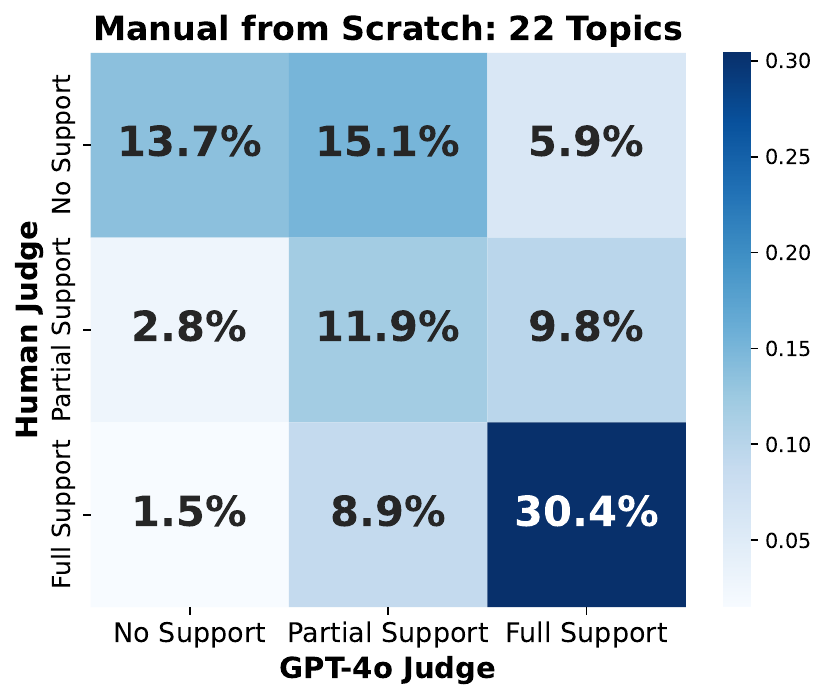}
        \hspace{1mm}
        \includegraphics[width=0.45\textwidth]{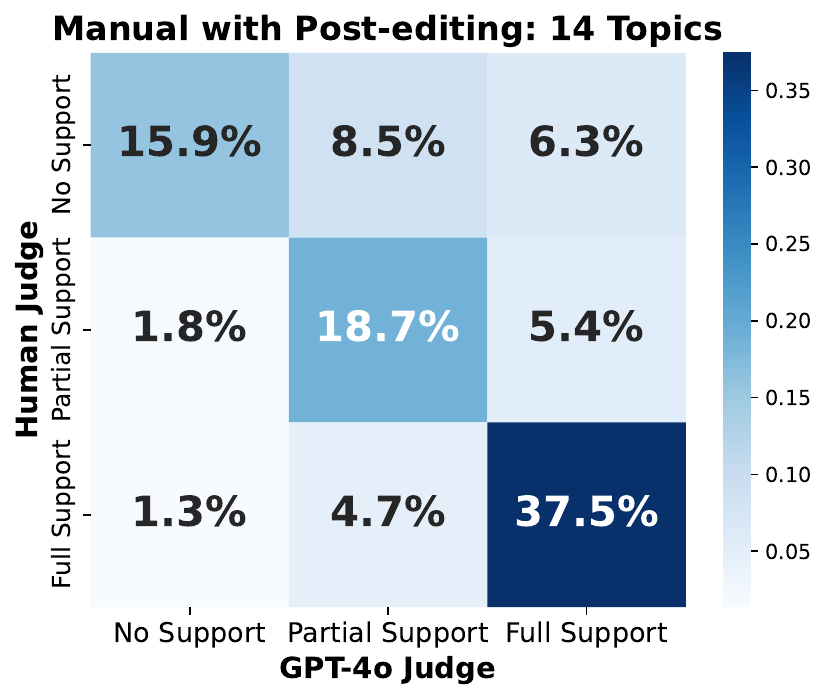}
    \caption{Confusion matrices comparing predictions from human and GPT-4o judges for the manual from-scratch condition (left) and the manual with post-editing condition (right).}
    \label{fig:conf_matrix}
\end{figure}

Next, to better understand how often the GPT-4o judge agrees with the human judges, we plot the confusion matrices in \autoref{fig:conf_matrix}.
We compare predictions by human judges with GPT-4o on two conditions:\ manual from scratch and manual with post-editing. 

\paragraph{\textbf{Manual from-scratch condition.}}
For 56\% (13.7\% + 11.9\% + 30.4\%), GPT-4o and the human judge perfectly agreed on their support judgment on 22 topics.
Both ``full support'' and ``no support'' categories have higher percentages (30.4\% and 13.7\%), showing that humans and GPT-4o as judges agreed more on both ends of the spectrum. 
For 15.1\%, the GPT-4o judge considered an annotation as ``partial support'', which the human judge annotated as ``no support''.
Finally, an important observation is that the GPT-4o judge is more likely to provide a higher support label than the human judge (the upper right triangle has a higher combined percentage over the lower left triangle).

\paragraph{\textbf{Manual with post-editing condition.}}
From the previous condition, we see the increase in perfect agreement rise to 72.1\% (15.9\% + 18.7\% + 37.5\%) on 14 topics that were annotated with post-editing GPT-4o labels. 
This shows that sentences and cited passages with ``partial support'' that led to disagreements in the manual from-scratch condition are reduced.
In this condition, human judges are more likely to agree with the GPT-4o judge unless it is an obvious mistake, i.e., when the GPT-4o judge considers an annotation to be ``full support'' and the human judge considers it to be ``no support'' (increased now to 6.3\% from 5.9\% in the manual from-scratch condition).

\begin{figure}
\begin{floatrow}
\ffigbox[\FBwidth]{%
  \resizebox{\linewidth}{!}{%
  \begin{tabular}{l r r r r }
    \toprule
    \multirow{2}{*}{Cohen's Kappa} & \multicolumn{2}{c}{From scratch} & \multicolumn{2}{c}{With post-editing} \\
    \cmidrule(lr){2-3} \cmidrule(lr){4-5}
    & GPT-4o & Human & GPT-4o & Human \\
    \midrule
    \midrule
    Independent human & \textbf{0.29} & $-$0.03 & \textbf{0.27} & 0.07 \\
    LLAMA-3.1 (405B) & \textbf{0.60} & $-$0.20 & \textbf{0.46} & $-$0.06 \\
  \bottomrule
  \end{tabular}}
  }{%
   \caption{Inter-annotator agreement score (Cohen's $\kappa$) for our unbiased study on disagreements between GPT-4o and human annotators.}\label{tab:inter-annotator}%
}
\ffigbox[\FBwidth]{%
\includegraphics[trim={0 10 0 10},width=0.95\linewidth]{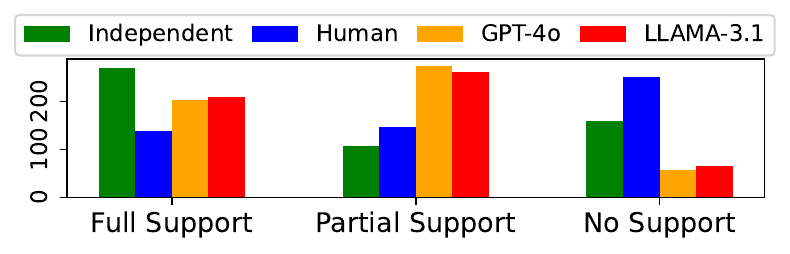}
}
{%
    \caption{Support label prediction by different judges for each support category (FS, PS, NS) in the disagreement analysis on 537 sentence--passage pairs.}\label{fig:label_distribution}\vspace{-1mm} 
  }%
\end{floatrow}
\vspace*{-\baselineskip}
\end{figure}

\section{Annotator Disagreements} 

In the experiments reported in Section~\ref{sec:results}, we observed frequent disagreements between the human and GPT-4o judge.
To further study this, we performed unbiased annotations from scratch by carefully re-assessing the support judgment of randomly sampled disagreements between the human and GPT-4o judge, with an independent human judge and another LLM judge using LLAMA-3.1 405B \cite{dubey2024llama} (with the same prompt as GPT-4o in \autoref{fig:prompt}). 
We randomly sampled 15 disagreement pairs per topic, re-evaluating 537 sentences and their first cited passages, including both assessment conditions:\ (1) manual from scratch and (2) manual with post-editing.

\paragraph{\textbf{Results.}}
As shown in \autoref{tab:inter-annotator}, we interestingly find the independent human judge to be \textit{better correlated} with GPT-4o than the human judge provided by NIST (Cohen's $\kappa$ of 0.29 and 0.27 versus $-$0.03 and 0.07) in the manual from-scratch condition. 
The independent judge fully matched 31\% of the time with the human judge and 51\% of the time with the GPT-4o judge.
Similarly, in the manual with post-editing condition, the independent judge fully matched 37\% of the time with the human judge and 52\% of the time with the GPT-4o judge.
LLAMA-3.1 405B had a stronger correlation with another LLM (GPT-4o) over human judges (Cohen's $\kappa$ of 0.60 and 0.46 versus $-$0.20 and $-$0.06), demonstrating the high likelihood of different LLMs providing similar prediction labels. 

From the label distributions in the disagreement analysis shown in \autoref{fig:label_distribution}, we observed that both LLMs (LLAMA-3.1 405B and GPT-4o) labeled about 49--51\% of sentences and their cited passage as ``partial support'', whereas the human judge labeled 47\% of the sentences as ``no support''. 
The independent judge labeled 50\% of the sentences as ``full support''. 
We keep as future work explorations of consistency:\ why LLM judges labeled only a few sentences as ``no support'', and similarly, why human judges labeled a majority of sentences as ``no support''.

\begin{table}[t]
\setlength{\tabcolsep}{3pt}

\renewcommand{\arraystretch}{1}
\centering
\resizebox{0.49\textwidth}{!}{
{\begin{tabular}{p{8cm}}
\toprule
{\bf Answer Sentence:} Swift started dating 19-year-old One Direction frontman, Harry Styles, barely after her breakup from Conor Kennedy. \\ \midrule
{\bf Passage ID [0]:} \texttt{doc\_04\_1081579649\#7\_2253255175} \\
{\bf Title:} A Timeline Of Taylor Swift's Age-Inappropriate Romances | Business Insider \\
{\bf Text:} Today, Taylor Swift turns 23, but you would never know how old the singer actually is based on her love life. From 18-year-old Conor Kennedy to 35-year-old John Mayer, Swift is no ageist when it comes to who she will date. Young or old, British or American, Swift really doesn’t discriminate in her love life. But at the age of 23, Swift has already had at least 10 high-profile relationships (some more flings, really) that have all ended the same way — in song. In honour of her 23rd year, let’s relive Swift’s vast array of boyfriends, shall we? [...] \\ [-0.5em] \textcolor{lightgray}{\rule[-0.1ex]{\linewidth}{0.08em}}
{\bf Human Judge:} \hlgreen{Full Support}  \textcolor{red}{$\boldsymbol{\times}$}  \\ 
{\bf GPT-4o Judge:} \hlred{No Support} \textcolor{dark_green}{$\boldsymbol{\checkmark}$} \\
\midrule
\midrule
{\bf Answer Sentence:} Her first Hollywood romance with Joe Jonas was age-appropriate, as both were 19 at the time. \\ \midrule
{\bf Passage ID [3]:} \texttt{doc\_04\_1081579649\#2\_2253244363}\\
{\bf Title:} Timeline of Taylor Swift's Relationships \\
{\bf Text:} [...] \hl{2008: Taylor Swift, 19 \& Joe Jonas, 19. Swift's first Hollywood romance was perfectly age-appropriate.} While Swift was just emerging onto the pop scene, Jonas, along with his singing and dancing brothers, were at the top of their game. [...] \\[-0.5em] \textcolor{lightgray}{\rule[-0.1ex]{\linewidth}{0.08em}}
{\bf Human Judge:} \hlred{No Support} \textcolor{red}{$\boldsymbol{\times}$} \\
{\bf GPT-4o Judge:} \hlgreen{Full Support} \textcolor{dark_green}{$\boldsymbol{\checkmark}$} \\
\bottomrule
\end{tabular}
}}
\hspace{0.1cm}
\resizebox{0.485\textwidth}{!}{
{\begin{tabular}{p{8cm}}
\toprule
{\bf Answer Sentence:} Swift has dated younger men, such as 17-year-old Taylor Lautner when she was 20, which was controversial due to the age difference. \\ \midrule
{\bf Passage ID [8]:} \texttt{doc\_35\_202251892\#7\_427547583} \\
{\bf Title:} A Timeline Of Taylor Swift's Age-Inappropriate Romances | Business Insider \\
{\bf Text:} 2009: \hl{Taylor Swift, 20, \& Taylor Lautner, 17. And here's where she started to slide. Technically, this relationship was possibly illegal in many states as 20-year-old Swift was dating a minor.} [...] In October 2010, Swift revealed her crush: `Taylor Lautner. It's always going to be Taylor.' It's believed that the song `Back To December' is Swift's apology to Lautner for their breakup. [...] \\[-0.5em] \textcolor{lightgray}{\rule[-0.1ex]{\linewidth}{0.08em}}
{\bf Human Judge:} \hlgreen{Full Support} \textcolor{dark_green}{$\boldsymbol{\checkmark}$} \\
{\bf GPT-4o Judge:} \hlorange{Partial Support} \textcolor{red}{$\boldsymbol{\times}$} \\
\midrule
\midrule
{\bf Answer Sentence:} Swift's youth and inexperience have been cited as factors in her relationships, with some partners reportedly taking advantage of her naivety. \\ \midrule
{\bf Passage ID [8]:} \texttt{doc\_48\_481103263\#4\_840591332} \\
{\bf Title:} Taylor Swift Boyfriends: Taylor Swift's Relationships | New Idea Magazine \\
{\bf Text:} [...] John Mayer (December 2009-February 2010)
\hl{The 11-year age gap between Taylor and John reportedly strained their relationship, with Taylor feeling that her naivety was taken advantage of.} She references their ugly breakup in her song “Dear John”. Songs: Dear John [...] \\ [-0.5em] \textcolor{lightgray}{\rule[-0.1ex]{\linewidth}{0.08em}}
{\bf Human Judge:} \hlgreen{Full Support} \textcolor{dark_green}{$\boldsymbol{\checkmark}$} \\
{\bf GPT-4o Judge:} \hlorange{Partial Support} \textcolor{red}{$\boldsymbol{\times}$} \\
\bottomrule
\end{tabular}}}
\caption{Examples of annotation mistakes by either GPT-4o or the human judge for the topic ``\textit{how taylor swift's age affects her relationships}'' taken from the disagreement analysis. The fragment of the passage that supports the answer sentence is \hl{highlighted}.}
\label{tab:annotation-mistakes}
\end{table}

\paragraph{\textbf{Qualitative analysis.}}
We further assessed examples qualitatively to understand failure cases, for example, when a human or GPT-4o judge makes mistakes during support evaluation.
In \autoref{tab:annotation-mistakes}, we show a few examples of annotation mistakes on the Taylor Swift topic found in either the human or GPT-4o judgment.
Overall, we summarize a few of the following errors made by GPT-4o:

\begin{itemize}[leftmargin=0.5cm]

\item GPT-4o can confuse words or phrases with similar meanings; for example, it is unable to distinguish between police and security specialists.

\item GPT-4o can miss out on evaluating the whole sentence (especially information present at the end of the sentence), biasing towards the ``full support'' label.

\item GPT-4o can label ``partial support'' if the theme in the answer sentence is similar, but the passage does not support any text present in the answer sentence, i.e., ``no support''.

\end{itemize}

On the other hand, human judges make mistakes due to not reading the passages carefully.
In some cases, answer sentences that were directly stated in the middle of or at the end of a passage, or mentioned in parts of the passage, were surprisingly unnoticed by a human judge. 
This causes the human judge to label such cases as ``no support'' instead of ``full support''.
Lastly, we observe that a human judge occasionally labels an answer sentence as ``full support'' even though the passage doesn't provide any support information.
We suspect that this could be due to an inherent bias relying on the human judge's memory or understanding of the topic, instead of strictly relying on the actual passage text.

\section{Conclusion}

In this work, we evaluated support in RAG answers by analyzing 45 submissions across 36 topics from the \trecrag in a large-scale comparative study involving both humans and LLMs as judges.
We critiqued and evaluated strong LLM judges, like GPT-4o, against human annotators for support assessment.

Our results show a high agreement between GPT-4o and human judgments, with a perfect match between judgments occurring 56\% of the time in the manual from-scratch condition, increasing to 72\% in the manual with post-editing condition. 
We observe that disagreements between humans and LLMs mainly occur for sentence--passage pairs indicating partial support, i.e., in the middle of the support evaluation spectrum. 

To better understand these disagreements, we conducted an unbiased evaluation by carefully re-assessing judgments with an independent human judge and a different LLM.
Interestingly, in cases of disagreements, both the independent human judge and the LLAMA-3.1 judge agreed more with the GPT-4o judge than with the human judge, providing evidence for widely divergent opinions and perhaps the veracity of using LLMs for support evaluation. 
Further research could explore the nuances of disagreements between human and LLM judges and investigate limitations of both humans and LLMs to improve future iterations of support assessment.

\section*{Acknowledgments}
This work would not have been possible without the annotator team at NIST.
We are grateful for their contributions.
This research was supported in part by the Natural Sciences and Engineering Research Council (NSERC) of Canada.
Additional funding is provided by Snowflake, Microsoft via the Accelerating Foundation Models Research program, and an Institute of Information \& Communications Technology Planning \& Evaluation (IITP) grant funded by the Korean Government (MSIT)\ (No.\ RS-2024-00457882, National AI Research Lab Project).
Thanks to Corby Rosset for providing the test queries, based on the methodology developed in Researchy Questions \cite{Rosset:2402.17896:2024}.

\bibliography{rag24-support}
\bibliographystyle{acl_natbib}




\begin{table}[p!]
\centering
\scalebox{0.68}{
\begin{tabular}{llrrrr}
\toprule
Run ID & Group & Task & Weighted Precision &  Weighted Recall & \#Sentences \\ \midrule
ag\_rag\_gpt35\_expansion\_rrf\_20 & IITD-IRL & RAG & 0.793 & 0.793 & 4.82 \\
Enhanced\_Iterative\_Fact\_Refinement\_and\_Prioritization & TREMA-UNH & RAG & 0.777 & 0.777 & 13.45 \\
UWCrag & WaterlooClarke & AG & 0.717 & 0.717 & 7.91 \\
Ranked\_Iterative\_Fact\_Extraction\_and\_Refinement & TREMA-UNH & RAG & 0.709 & 0.709 & 13.68 \\
ldilab\_gpt\_4o & ldisnu & RAG & 0.698 & 0.445 & 12.86 \\
zeph\_test\_rag\_rrf\_expand\_query & IITD-IRL & AG & 0.681 & 0.681 & 4.64 \\
dilab\_repllama\_listt5\_pass3\_gpt4o & ldisnu & AG & 0.669 & 0.408 & 13.45 \\
baseline\_frag\_rag24.test\_gpt-4o\_top20 & coordinators & AG & 0.654 & 0.401 & 14.18 \\
cir\_gpt-4o-mini\_no\_reranking\_50\_0.5\_100\_301\_p1 & CIR & RAG & 0.646 & 0.646 & 7.05 \\
neurag & neu & AG & 0.644 & 0.397 & 15.82 \\
baseline\_frag\_rag24.test\_command-r-plus\_top20 & coordinators & AG & 0.642 & 0.583 & 11.68 \\
iiia\_dedup\_p1\_straight\_ag & IIIA-UNIPD & RAG & 0.635 & 0.572 & 3.77 \\
listgalore\_gpt4o\_ragnarokv4\_top20 & h2oloo & AG & 0.610 & 0.487 & 12.14 \\
listgalore\_l31-70b\_ragnarokv4\_top20 & h2oloo & AG & 0.605 & 0.532 & 9.64 \\
cohere+post\_processing & KML & RAG & 0.582 & 0.542 & 18.0 \\
neuragfix & neu & AG & 0.569 & 0.359 & 15.82 \\
baseline\_rag24.test\_l31\_70b\_instruct\_top20 & coordinators & RAG & 0.568 & 0.556 & 7.5 \\
ielab-b70bf-70bqp-70bafs & ielab & AG & 0.565 & 0.445 & 3.14 \\
iiia\_standard\_p1\_straight\_ag & IIIA-UNIPD & RAG & 0.562 & 0.483 & 6.23 \\
FT-llama3 & uog-tht & AG & 0.553 & 0.553 & 4.64 \\
UDInfolab.RAG.Query & InfoLab & AG & 0.534 & 0.534 & 6.86 \\
webis-rag-run0-taskrag & webis & AG & 0.527 & 0.527 & 6.27 \\
baseline\_top\_5 & uis-iai & RAG & 0.522 & 0.522 & 3.23 \\
agtask-bm25-colbert\_faiss-gpt4o-llama70b & softbank-meisei & RAG & 0.516 & 0.516 & 8.27 \\
UDInfolab.RAG.AnsAI & InfoLab & AG & 0.505 & 0.505 & 7.18 \\
cir\_gpt-4o-mini\_Cosine\_50\_0.5\_100\_301\_p1 & CIR & RAG & 0.495 & 0.495 & 7.05 \\
buw & buw & AG & 0.481 & 0.481 & 8.14 \\
webis-rag-run1-taskrag & webis & AG & 0.464 & 0.464 & 6.18 \\
oneshot\_post\_sentenced & buw & AG & 0.461 & 0.356 & 9.09 \\
rag\_bm25-colbert\_faiss-gpt4o-llama70b & softbank-meisei & AG & 0.433 & 0.433 & 7.27 \\
ruc001 & Ruc01 & AG & 0.425 & 0.348 & 14.77 \\
gpt\_mini & KML & RAG & 0.368 & 0.368 & 7.45 \\
ginger\_top\_5 & uis-iai & RAG & 0.367 & 0.367 & 4.32 \\
LAS-splade-mxbai-mmr8-RAG & ncsu-las & AG & 0.308 & 0.226 & 12.82 \\
UWCgarag & WaterlooClarke & AG & 0.305 & 0.289 & 9.95 \\
iiresearch-bm25-top10-llama3-8b-instruct & ii\_research & AG & 0.293 & 0.293 & 4.05 \\
BEST\_cot\_gpt3.5 & citi & AG & 0.291 & 0.256 & 6.23 \\
ICL-mistral & uog-tht & AG & 0.267 & 0.201 & 5.82 \\
ielab-b70bf-70bqfs-ad\_hoc & ielab & AG & 0.246 & 0.246 & 4.59 \\
SECOND\_cot\_gpt3.5 & citi & AG & 0.246 & 0.192 & 5.36 \\
ISIR-IRIT-zephyr\_query\_gen & IRIT & AG & 0.205 & 0.115 & 2.59 \\
LAS-splade-mxbai-rrf-mmr8 & ncsu-las & AG & 0.167 & 0.116 & 13.41 \\
ISIR-IRIT-zephyr\_p2 & IRIT & AG & 0.131 & 0.094 & 1.59 \\
qrant\_bge\_gemini & SGU & AG & 0.130 & 0.130 & 6.45 \\
webis-manual & webis & AG & 0.079 & 0.037 & 1.68 \\
\bottomrule
\end{tabular}}
\caption{
Weighted precision and recall scores for the top two runs from each group in \trecrag under the \textbf{manual from-scratch} condition on 22 topics evaluated by \textbf{human judges} provided by NIST. \#Sentences denotes the average number of sentences in the participant's submitted answer.}
\label{tab:support_manual_from_scratch}
\end{table}
\begin{table}[p!]
\centering
\scalebox{0.68}{
\begin{tabular}{llrrrr}
\toprule
Run ID & Group & Task & Weighted Precision &  Weighted Recall & \#Sentences \\ \midrule
ag\_rag\_gpt35\_expansion\_rrf\_20 & IITD-IRL & RAG & 0.868 & 0.868 & 4.82 \\
Ranked\_Iterative\_Fact\_Extraction\_and\_Refinement & TREMA-UNH & RAG & 0.861 & 0.861 & 13.68 \\
Enhanced\_Iterative\_Fact\_Refinement\_and\_Prioritization & TREMA-UNH & RAG & 0.846 & 0.846 & 13.45 \\
UWCrag & WaterlooClarke & AG & 0.841 & 0.841 & 7.91 \\
zeph\_test\_rag\_rrf\_expand\_query & IITD-IRL & AG & 0.807 & 0.807 & 4.64 \\
ldilab\_gpt\_4o & ldisnu & RAG & 0.791 & 0.514 & 12.86 \\
dilab\_repllama\_listt5\_pass3\_gpt4o & ldisnu & AG & 0.784 & 0.491 & 13.45 \\
baseline\_frag\_rag24.test\_command-r-plus\_top20 & coordinators & AG & 0.778 & 0.718 & 11.68 \\
iiia\_dedup\_p1\_straight\_ag & IIIA-UNIPD & RAG & 0.767 & 0.715 & 3.77 \\
cir\_gpt-4o-mini\_no\_reranking\_50\_0.5\_100\_301\_p1 & CIR & RAG & 0.757 & 0.757 & 7.05 \\
baseline\_frag\_rag24.test\_gpt-4o\_top20 & coordinators & AG & 0.741 & 0.467 & 14.18 \\
neuragfix & neu & AG & 0.737 & 0.459 & 15.82 \\
neurag & neu & AG & 0.735 & 0.458 & 15.82 \\
ielab-b70bf-70bqp-70bafs & ielab & AG & 0.733 & 0.591 & 3.18 \\
baseline\_rag24.test\_l31\_70b\_instruct\_top20 & coordinators & RAG & 0.731 & 0.719 & 7.50 \\
cohere+post\_processing & KML & RAG & 0.712 & 0.671 & 18.00 \\
listgalore\_l31-70b\_ragnarokv4\_top20 & h2oloo & AG & 0.711 & 0.622 & 9.64 \\
iiia\_standard\_p1\_straight\_ag & IIIA-UNIPD & RAG & 0.702 & 0.610 & 6.23 \\
webis-manual & webis & AG & 0.702 & 0.415 & 14.14 \\
listgalore\_gpt4o\_ragnarokv4\_top20 & h2oloo & AG & 0.699 & 0.563 & 12.14 \\
cir\_gpt-4o-mini\_Cosine\_50\_0.5\_100\_301\_p1 & CIR & RAG & 0.677 & 0.677 & 7.05 \\
UDInfolab.RAG.AnsAI & InfoLab & AG & 0.659 & 0.659 & 7.18 \\
FT-llama3 & uog-tht & AG & 0.640 & 0.640 & 4.64 \\
baseline\_top\_5 & uis-iai & RAG & 0.639 & 0.639 & 3.23 \\
agtask-bm25-colbert\_faiss-gpt4o-llama70b & softbank-meisei & RAG & 0.633 & 0.633 & 8.27 \\
UDInfolab.RAG.Query & InfoLab & AG & 0.623 & 0.623 & 6.86 \\
webis-rag-run0-taskrag & webis & AG & 0.616 & 0.616 & 6.27 \\
buw & buw & AG & 0.597 & 0.597 & 8.14 \\
oneshot\_post\_sentenced & buw & AG & 0.581 & 0.445 & 9.09 \\
ruc001 & Ruc01 & AG & 0.577 & 0.479 & 14.77 \\
webis-rag-run1-taskrag & webis & AG & 0.531 & 0.531 & 6.18 \\
gpt\_mini & KML & RAG & 0.494 & 0.494 & 7.45 \\
rag\_bm25-colbert\_faiss-gpt4o-llama70b & softbank-meisei & AG & 0.475 & 0.475 & 7.27 \\
ICL-mistral & uog-tht & AG & 0.454 & 0.345 & 5.82 \\
UWCgarag & WaterlooClarke & AG & 0.443 & 0.425 & 9.95 \\
iiresearch-bm25-top10-llama3-8b-instruct & ii\_research & AG & 0.433 & 0.392 & 4.14 \\
ginger\_top\_5 & uis-iai & RAG & 0.419 & 0.419 & 4.32 \\
LAS-splade-mxbai-mmr8-RAG & ncsu-las & AG & 0.376 & 0.270 & 12.82 \\
BEST\_cot\_gpt3.5 & citi & AG & 0.329 & 0.274 & 6.45 \\
SECOND\_cot\_gpt3.5 & citi & AG & 0.319 & 0.229 & 5.50 \\
ISIR-IRIT-zephyr\_query\_gen & IRIT & AG & 0.273 & 0.132 & 6.95 \\
ielab-b70bf-70bqfs-ad\_hoc & ielab & AG & 0.253 & 0.253 & 4.59 \\
LAS-splade-mxbai-rrf-mmr8 & ncsu-las & AG & 0.237 & 0.170 & 13.41 \\
ISIR-IRIT-zephyr\_p2 & IRIT & AG & 0.159 & 0.112 & 6.73 \\
qrant\_bge\_gemini & SGU & AG & 0.157 & 0.157 & 6.45 \\
\bottomrule
\end{tabular}}
\caption{
Weighted precision and recall scores for the top two runs from each group in \trecrag under the \textbf{manual from-scratch} condition on 22 topics evaluated by the \textbf{GPT-4o judge}. \#Sentences denotes the average number of sentences in the participant's submitted answer.}
\label{tab:support_manual_from_scratch_auto}
\end{table}

\begin{table}[p]
\centering
\scalebox{0.68}{
\begin{tabular}{llrrrr}
\toprule
Run ID & Group & Task & Weighted Precision &  Weighted Recall & \#Sentences \\ \midrule
ag\_rag\_gpt35\_expansion\_rrf\_20 & IITD-IRL & RAG & 0.793 & 0.793 & 4.79 \\
Enhanced\_Iterative\_Fact\_Refinement\_and\_Prioritization & TREMA-UNH & RAG & 0.773 & 0.773 & 14.5 \\
Ranked\_Iterative\_Fact\_Extraction\_and\_Refinement & TREMA-UNH & RAG & 0.772 & 0.772 & 14.07 \\
baseline\_frag\_rag24.test\_command-r-plus\_top20 & coordinators & AG & 0.760 & 0.683 & 12.29 \\
neurag & neu & AG & 0.735 & 0.442 & 15.71 \\
listgalore\_gpt4o\_ragnarokv4\_top20 & h2oloo & AG & 0.734 & 0.575 & 12.21 \\
baseline\_frag\_rag24.test\_gpt-4o\_top20 & coordinators & AG & 0.726 & 0.404 & 14.64 \\
UWCrag & WaterlooClarke & AG & 0.724 & 0.724 & 6.64 \\
zeph\_test\_rag\_rrf\_expand\_query & IITD-IRL & AG & 0.709 & 0.709 & 4.07 \\
dilab\_repllama\_listt5\_pass3\_gpt4o & ldisnu & AG & 0.709 & 0.418 & 12.0 \\
listgalore\_l31-70b\_ragnarokv4\_top20 & h2oloo & AG & 0.679 & 0.593 & 9.57 \\
baseline\_rag24.test\_l31\_70b\_instruct\_top20 & coordinators & RAG & 0.668 & 0.660 & 7.5 \\
neuragfix & neu & AG & 0.666 & 0.398 & 15.71 \\
iiia\_standard\_p1\_straight\_ag & IIIA-UNIPD & RAG & 0.662 & 0.602 & 4.93 \\
ielab-b70bf-70bqp-70bafs & ielab & AG & 0.658 & 0.571 & 2.86 \\
UDInfolab.RAG.Query & InfoLab & AG & 0.654 & 0.654 & 7.71 \\
iiia\_dedup\_p1\_straight\_ag & IIIA-UNIPD & RAG & 0.650 & 0.634 & 4.43 \\
ldilab\_gpt\_4o & ldisnu & RAG & 0.646 & 0.363 & 12.79 \\
cir\_gpt-4o-mini\_no\_reranking\_50\_0.5\_100\_301\_p1 & CIR & RAG & 0.644 & 0.644 & 6.86 \\
UDInfolab.RAG.AnsAI & InfoLab & AG & 0.639 & 0.639 & 7.43 \\
webis-rag-run0-taskrag & webis & AG & 0.611 & 0.611 & 5.21 \\
baseline\_top\_5 & uis-iai & RAG & 0.571 & 0.571 & 3.07 \\
FT-llama3 & uog-tht & AG & 0.557 & 0.557 & 3.29 \\
cohere+post\_processing & KML & RAG & 0.556 & 0.556 & 18.79 \\
cir\_gpt-4o-mini\_Cosine\_50\_0.5\_100\_301\_p1 & CIR & RAG & 0.538 & 0.538 & 7.29 \\
agtask-bm25-colbert\_faiss-gpt4o-llama70b & softbank-meisei & RAG & 0.526 & 0.526 & 7.43 \\
ruc001 & Ruc01 & AG & 0.516 & 0.458 & 11.21 \\
oneshot\_post\_sentenced & buw & AG & 0.460 & 0.357 & 12.0 \\
buw & buw & AG & 0.459 & 0.459 & 9.5 \\
gpt\_mini & KML & RAG & 0.417 & 0.417 & 8.36 \\
ginger\_top\_5 & uis-iai & RAG & 0.413 & 0.413 & 3.79 \\
rag\_bm25-colbert\_faiss-gpt4o-llama70b & softbank-meisei & AG & 0.398 & 0.398 & 7.29 \\
webis-rag-run1-taskrag & webis & AG & 0.356 & 0.356 & 5.57 \\
ielab-b70bf-70bqfs-ad\_hoc & ielab & AG & 0.352 & 0.350 & 4.71 \\
LAS-splade-mxbai-mmr8-RAG & ncsu-las & AG & 0.322 & 0.241 & 11.71 \\
iiresearch-bm25-top10-llama3-8b-instruct & ii\_research & AG & 0.321 & 0.321 & 1.86 \\
SECOND\_cot\_gpt3.5 & citi & AG & 0.287 & 0.224 & 5.14 \\
ICL-mistral & uog-tht & AG & 0.250 & 0.196 & 4.29 \\
UWCgarag & WaterlooClarke & AG & 0.237 & 0.201 & 9.29 \\
BEST\_cot\_gpt3.5 & citi & AG & 0.216 & 0.170 & 3.71 \\
LAS-splade-mxbai-rrf-mmr8 & ncsu-las & AG & 0.190 & 0.144 & 12.21 \\
qrant\_bge\_gemini & SGU & AG & 0.147 & 0.147 & 5.79 \\
ISIR-IRIT-zephyr\_p2 & IRIT & AG & 0.143 & 0.047 & 2.86 \\
ISIR-IRIT-zephyr\_query\_gen & IRIT & AG & 0.143 & 0.055 & 1.79 \\
webis-manual & webis & AG & 0.106 & 0.075 & 1.71 \\
\bottomrule
\end{tabular}}
\caption{
Weighted precision and recall scores for the top two runs from each group in \trecrag under the \textbf{manual with post-editing} condition on 14 topics evaluated by \textbf{human judges} provided by NIST. \#Sentences denotes the average number of sentences in the participant's submitted answer.}
\label{tab:support_manual_with_post_editing}
\end{table}
\begin{table}[p]
\centering
\scalebox{0.68}{
\begin{tabular}{llrrrr}
\toprule
Run ID & Group & Task & Weighted Precision &  Weighted Recall & \#Sentences \\ \midrule
Ranked\_Iterative\_Fact\_Extraction\_and\_Refinement & TREMA-UNH & RAG & 0.900 & 0.900 & 14.07 \\
Enhanced\_Iterative\_Fact\_Refinement\_and\_Prioritization & TREMA-UNH & RAG & 0.885 & 0.885 & 14.5 \\
zeph\_test\_rag\_rrf\_expand\_query & IITD-IRL & AG & 0.852 & 0.852 & 4.07 \\
ldilab\_gpt\_4o & ldisnu & RAG & 0.842 & 0.482 & 12.79 \\
ag\_rag\_gpt35\_expansion\_rrf\_20 & IITD-IRL & RAG & 0.838 & 0.838 & 4.79 \\
dilab\_repllama\_listt5\_pass3\_gpt4o & ldisnu & AG & 0.831 & 0.474 & 12.0 \\
UWCrag & WaterlooClarke & AG & 0.817 & 0.817 & 6.64 \\
listgalore\_l31-70b\_ragnarokv4\_top20 & h2oloo & AG & 0.807 & 0.687 & 9.57 \\
baseline\_frag\_rag24.test\_command-r-plus\_top20 & coordinators & AG & 0.805 & 0.724 & 12.29 \\
iiia\_dedup\_p1\_straight\_ag & IIIA-UNIPD & RAG & 0.783 & 0.740 & 4.43 \\
baseline\_frag\_rag24.test\_gpt-4o\_top20 & coordinators & AG & 0.782 & 0.427 & 14.64 \\
neuragfix & neu & AG & 0.775 & 0.464 & 15.71 \\
listgalore\_gpt4o\_ragnarokv4\_top20 & h2oloo & AG & 0.775 & 0.606 & 12.21 \\
cir\_gpt-4o-mini\_Cosine\_50\_0.5\_100\_301\_p1 & CIR & RAG & 0.773 & 0.773 & 7.29 \\
webis-manual & webis & AG & 0.770 & 0.436 & 12.64 \\
neurag & neu & AG & 0.762 & 0.455 & 15.71 \\
cir\_gpt-4o-mini\_no\_reranking\_50\_0.5\_100\_301\_p1 & CIR & RAG & 0.750 & 0.750 & 6.86 \\
UDInfolab.RAG.Query & InfoLab & AG & 0.731 & 0.731 & 7.71 \\
baseline\_rag24.test\_l31\_70b\_instruct\_top20 & coordinators & RAG & 0.728 & 0.719 & 7.5 \\
ielab-b70bf-70bqp-70bafs & ielab & AG & 0.723 & 0.620 & 2.93 \\
UDInfolab.RAG.AnsAI & InfoLab & AG & 0.717 & 0.717 & 7.43 \\
iiia\_standard\_p1\_straight\_ag & IIIA-UNIPD & RAG & 0.690 & 0.621 & 4.93 \\
agtask-bm25-colbert\_faiss-gpt4o-llama70b & softbank-meisei & RAG & 0.687 & 0.687 & 7.43 \\
webis-rag-run0-taskrag & webis & AG & 0.682 & 0.682 & 5.21 \\
cohere+post\_processing & KML & RAG & 0.682 & 0.682 & 18.79 \\
baseline\_top\_5 & uis-iai & RAG & 0.601 & 0.601 & 3.07 \\
FT-llama3 & uog-tht & AG & 0.574 & 0.574 & 3.29 \\
oneshot\_post\_sentenced & buw & AG & 0.571 & 0.442 & 12.0 \\
buw & buw & AG & 0.558 & 0.558 & 9.5 \\
ruc001 & Ruc01 & AG & 0.548 & 0.482 & 11.29 \\
webis-rag-run1-taskrag & webis & AG & 0.487 & 0.487 & 5.57 \\
gpt\_mini & KML & RAG & 0.471 & 0.440 & 8.36 \\
rag\_bm25-colbert\_faiss-gpt4o-llama70b & softbank-meisei & AG & 0.435 & 0.435 & 7.29 \\
ginger\_top\_5 & uis-iai & RAG & 0.408 & 0.408 & 3.79 \\
UWCgarag & WaterlooClarke & AG & 0.387 & 0.324 & 9.29 \\
SECOND\_cot\_gpt3.5 & citi & AG & 0.369 & 0.297 & 5.5 \\
ICL-mistral & uog-tht & AG & 0.357 & 0.268 & 4.29 \\
LAS-splade-mxbai-mmr8-RAG & ncsu-las & AG & 0.347 & 0.261 & 11.71 \\
ielab-b70bf-70bqfs-ad\_hoc & ielab & AG & 0.297 & 0.295 & 4.71 \\
iiresearch-bm25-top10-llama3-8b-instruct & ii\_research & AG & 0.296 & 0.296 & 1.86 \\
ISIR-IRIT-zephyr\_p2 & IRIT & AG & 0.243 & 0.074 & 6.36 \\
BEST\_cot\_gpt3.5 & citi & AG & 0.227 & 0.161 & 5.5 \\
LAS-splade-mxbai-rrf-mmr8 & ncsu-las & AG & 0.195 & 0.150 & 12.21 \\
ISIR-IRIT-zephyr\_query\_gen & IRIT & AG & 0.179 & 0.053 & 5.36 \\
qrant\_bge\_gemini & SGU & AG & 0.149 & 0.149 & 5.79 \\
\bottomrule
\end{tabular}}
\caption{Weighted precision and recall scores for the top two runs from each group in \trecrag under the \textbf{manual with post-editing} condition on 14 topics evaluated by the \textbf{GPT-4o judge}. \#Sentences denotes the average number of sentences in the participant's submitted answer.}
\label{tab:support_manual_with_post_editing_auto}
\end{table}


\end{document}